\pdfoutput=1

\documentclass[11pt]{article}

\usepackage[final]{ACL2023}

\usepackage{times}
\usepackage{latexsym}

\usepackage[T1]{fontenc}

\usepackage[utf8]{inputenc}

\usepackage{microtype}

\usepackage{hyperref}
\usepackage{url}
\usepackage{graphicx}

\usepackage{algorithm}
\usepackage{algorithmic}

\usepackage{booktabs}
\usepackage{float}
\usepackage{amsmath}
\usepackage{amsfonts,amssymb}
\usepackage{mathrsfs}
\usepackage{multirow}
\usepackage{makecell}

\usepackage{pgfplots}
\usepackage{newfloat}
\usepackage{listings}
\usepackage{subfigure}
\usepackage{enumitem}
\usepackage{amssymb}

\usepackage{pgfplots}

\definecolor{win}{RGB}{76,153,0}
\definecolor{tie}{RGB}{51,153,255}
\definecolor{lose}{RGB}{255,80,80}

\setenumerate[1]{itemsep=2pt,partopsep=0pt,parsep=\parskip,topsep=2pt}
\setitemize[1]{itemsep=2pt,partopsep=0pt,parsep =\parskip,topsep=2pt}
\setdescription{itemsep=2pt,partopsep=0pt,parsep=\parskip,topsep=2pt}

\title{\includegraphics[width=0.6cm]{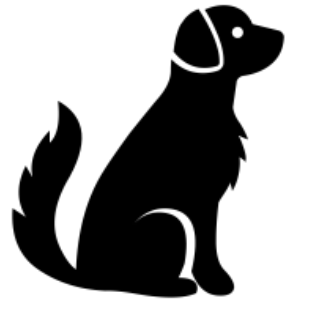}DoG-Instruct: Towards Premium Instruction-Tuning Data via Text-Grounded Instruction Wrapping}

\author{
Yongrui Chen\textsuperscript{\rm 1,2}, 
Haiyun Jiang\textsuperscript{\rm 3,\thanks{\ \ Corresponding Author}}, 
Xinting Huang\textsuperscript{\rm 3}, 
Shuming Shi\textsuperscript{\rm 3} 
\& Guilin Qi\textsuperscript{\rm 1,2\thanks{\ \ Corresponding Author}}  \\
\textsuperscript{\rm 1}Southeast University \\
\textsuperscript{\rm 2}Key Laboratory of New Generation Artificial Intelligence Technology \\and Its Interdisciplinary Applications\ (Southeast University), Ministry of Education \\
\texttt{\{yrchen,gqi\}@seu.edu.cn} \\
\textsuperscript{\rm 3}Tencent AI Lab \\
\texttt{\{haiyunjiang,jeffjhhuang,shumingshi\}@tencent.com} \\
}

\begin{document}
\maketitle
\begin{abstract}
The improvement of LLMs' instruction-following capabilities relies heavily on the availability of high-quality instruction-response pairs. 
Unfortunately, the current methods used to collect the pairs suffer from either unaffordable labor costs or severe hallucinations in the self-generation of LLM.
To tackle these challenges, this paper proposes a scalable solution.
It involves training LLMs to generate instruction-response pairs based on human-written documents, rather than relying solely on self-generation without context.
Our proposed method not only exploits the advantages of human-written documents in reducing hallucinations but also utilizes an LLM to wrap the expression of documents, which enables us to bridge the gap between various document styles and the standard AI response.
Experiments demonstrate that our method outperforms existing typical methods on multiple benchmarks.
In particular, compared to the best-performing baseline, the LLM trained using our generated dataset exhibits a 10\% relative improvement in performance on AlpacaEval, despite utilizing only 1/5 of its training data.
Furthermore, a comprehensive manual evaluation validates the quality of the data we generated.
Our trained wrapper is publicly available\footnote[1]{\url{https://github.com/Bahuia/Dog-Instruct}}.
\end{abstract}

\section{Introduction\label{sec:intro}}


Recent efforts in the NLP community have focused on \textit{instruction-tuning}~\citep{DBLP:conf/iclr/SanhWRBSACSRDBX22,DBLP:conf/acl/MishraKBH22,DBLP:conf/iclr/WeiBZGYLDDL22}, i.e., improving large language models' (LLMs) capacity to understand and follow instructions~\citep{DBLP:conf/nips/BrownMRSKDNSSAA20,DBLP:journals/corr/abs-2204-02311,DBLP:journals/corr/abs-2302-13971}. 
Advanced LLMs have been trained to be capable of generating customized outputs when provided with specific instructions (with inputs), enabling them to adapt to new tasks without prior exposure.

As a crucial problem in improving LLMs' instruction-following capability, how to collect high-quality instruction-response pairs is gaining popularity. 
The majority of existing methods either rely on hiring professionals to write instructions for various NLP tasks~\citep{DBLP:conf/emnlp/WangMAKMNADASPK22,DatabricksBlog2023DollyV2} or promote the use of LLMs to automatically generate instructions~\citep{DBLP:conf/acl/WangKMLSKH23,alpaca,DBLP:journals/corr/abs-2305-14327}.
Unfortunately, these methods have limitations either in terms of scalability due to the labor-intensive nature of the annotation process or in terms of data quality due to the hallucination problem~\citep{DBLP:journals/corr/abs-2305-13534,zheng2023does} associated with LLMs.

Recent research~\citep{DBLP:journals/corr/abs-2304-08460,DBLP:journals/corr/abs-2308-06259} has provided a more potential idea: first directly using human-written documents as typical responses and then utilizing LLMs to predict the latent user instructions. 
This method, known as \textit{instruction back-translation}~\citep{DBLP:journals/corr/abs-2308-06259}, is based on the belief that human-written documents are inherently less prone to hallucinations compared to responses generated solely by LLMs.

\begin{figure*}
\centering
	\includegraphics[width=1\textwidth]{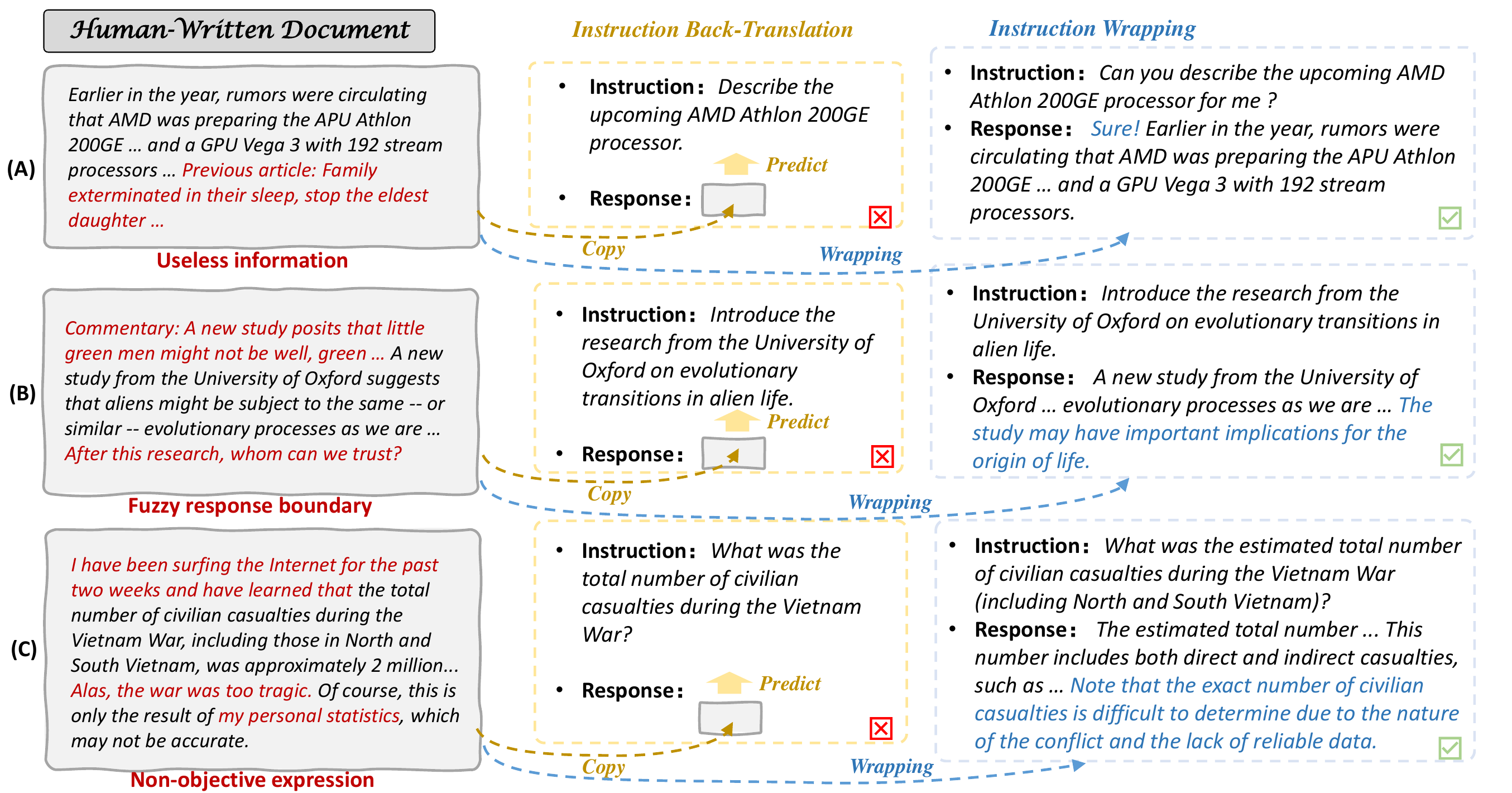}
	\caption{Differences between our proposed \textit{instruction wrapping} with \textit{instruction back-translation}~\citep{DBLP:journals/corr/abs-2304-08460,DBLP:journals/corr/abs-2308-06259}. Red text is not appropriate for responses. Blue text indicates that the original text has been added, deleted, or rewritten by LLM to align more closely with the desired standardized response.} \label{fig:example}
\end{figure*}

However, we argue that even if a document is free of hallucinations, it is not always appropriate to employ it directly as a typical response.
This is attributed to two main reasons:
a) First, not all parts of a document are valuable in constructing a response. 
For example, the red part of the document (\texttt{A}) in Figure \ref{fig:example} is completely useless for back-translating the resulting instruction (the gold box).
Moreover, valuable parts of the document often have fuzzy boundaries. 
For instance, the red text of (\texttt{B}) aims to create a tense atmosphere, again unsuitable to keep in response, but it also has some relevance to the topic (alien research) and is therefore difficult to be filtered out by simple preprocessing. 
b) Second, due to the different purposes of writing, there are often gaps in expression between the raw documents and the standard responses. 
As an illustration, the red portion of (\texttt{C}) contains multiple subjective descriptions, which deviates from the expected objectivity of an AI assistant.

In this paper, we propose a new paradigm for constructing instruction-tuned data, called \textit{instruction wrapping}.
It aims to train an open-sourced LLM to identify valuable parts from the original document and further transform them into fluent and objective instruction-response pairs.

\begin{figure*}
\centering
	\includegraphics[width=1\textwidth]{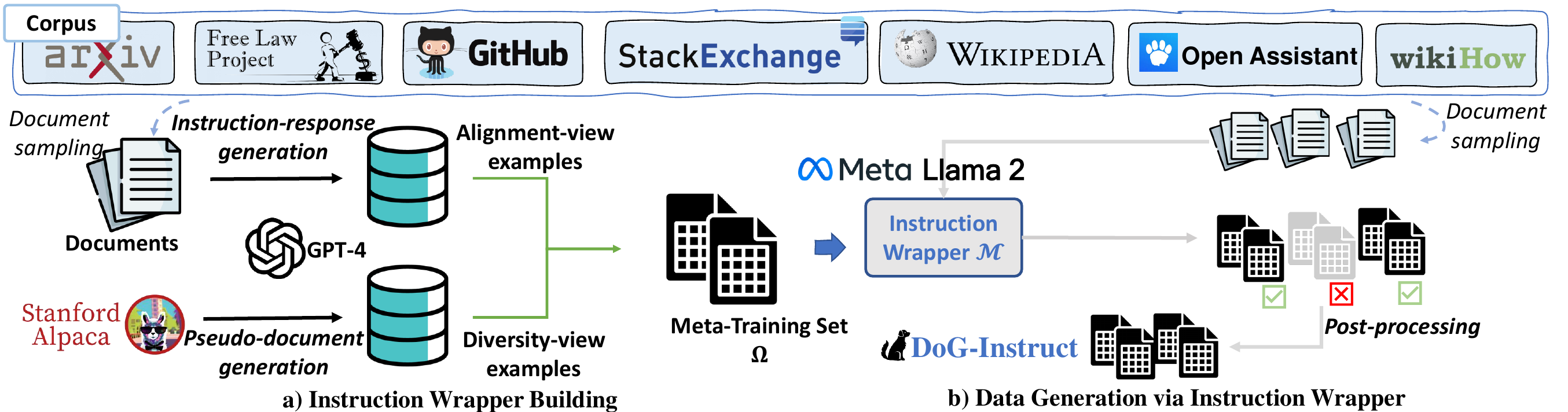}
	\caption{Overview of \textsc{DoG-Instruct} construction process. In stage a), a meta-training set $\Omega$ is constructed using GPT-4 and utilized to train the instruction wrapper. In stage b), the wrapper generates instruction-response pairs for each sampled document, and a post-processing strategy is employed to filter out invalid examples.} \label{fig:overview}
\end{figure*}

Briefly, our proposed method consists of two stages as shown in Figure \ref{fig:overview}.
In stage a), a well-aligned LLM is employed as the teacher to construct a meta-training set $\Omega$ for instruction wrapping. 
Each example in this set comprises a sampled document and its corresponding instruction-response pair, involving one of the following two views.
In the alignment view, we employ in-context learning to guide the teacher LLM in generating instruction-response pairs based on human-written documents. 
It allows for the adaptation of the teacher LLM to various real document styles.
In the diversity view, we begin with an existing diverse instruction set and prompt the teacher LLM to reversely generate a pseudo-document for each instruction-response pair. 
It ensures the training examples maintain instruction diversity.
Subsequently, we use the meta-training set to perform supervised fine-tuning on a publicly released LLM, which serves as our instruction wrapper.
In stage b), human-written documents from multiple domains are fed into our trained wrapper to generate instruction-response pairs. 
Then, a simple but efficient post-processing strategy is adopted to filter invalid examples based on the literal similarity.
Eventually, we name the resulting dataset Document-Grounded Instructions (\includegraphics[width=0.35cm]{figure/logo.pdf}\textsc{DoG-Instruct}), containing 12.4K instruction-response pairs. 

The LLM trained using \textsc{DoG-Instruct} achieves a remarkable 4.8\% improvement in performance on AlpacaEval compared to the best-performing baseline, while using only 1/5 of the training data. 
Furthermore, it achieves state-of-the-art results on the other three widely-used benchmarks.
Through further manual evaluation,  we illustrate that our \textsc{DoG-Instruct} method effectively mitigates the issue of hallucination while aligning the raw document with the desired response in terms of style.
In summary, the contributions of this paper include:
\begin{itemize}[leftmargin=1em]
    \item We propose a novel paradigm that trains LLMs to generate instruction-response pairs based on human-written documents. It not only leverages the document to reduce the hallucinations of responses, but also aligns the style of the raw document with the ideal response using LLM.

    \item We release a well-trained \textsc{Llama}-based instruction wrapper capable of consistently generating high-quality instruction-response pairs for documents across multiple domains.
    
    \item We conducted a comprehensive evaluation, both automatic and manual, which demonstrates that the LLM trained using our generated data outperforms all the compared baselines.
\end{itemize}

\section{Problem Formulation}
Given a set of documents $\{\mathcal{D}_1, \mathcal{D}_2, ..., \mathcal{D}_n\}$, where each $\mathcal{D}_i$ is a human-written document, our goal is to construct a set of pairs $\{(\mathcal{X}_1, \mathcal{Y}_1), ..., (\mathcal{X}_m, \mathcal{Y}_m)\}$, where $m \le n$, $\mathcal{X}_i$ and $\mathcal{Y}_i$ denote the instruction and response, respectively, and $(\mathcal{X}_i, \mathcal{Y}_i) := \mathcal{M}(\mathcal{D}_j)$. 
Here $\mathcal{M}$ is an LLM-based instruction wrapper that transforms $\mathcal{D}_i$ into an instruction-response pair.

\section{Collection of DoG-Instruct Data}
Figure~\ref{fig:overview} shows the entire process of our method. 
a) First, the instruction wrapper $\mathcal{M}$ is trained using the meta-training set $\Omega$, which is constructed by the well-aligned GPT-4. 
b) Subsequently, $\mathcal{M}$ takes sampled documents $\mathcal{D}$ as inputs to generate $(\mathcal{X}, \mathcal{Y})$ for constructing \textsc{DoG-Instruct}. 

\subsection{Corpus \& Document Sampling}
\label{sec:corpus}
To create a diverse set of documents, we utilize the Pile~\citep{DBLP:journals/corr/abs-2101-00027} corpus, which is a multi-domain collection of human-written documents. 
From the Pile, we sample documents from six different domains: ArXiv, FreeLaw, StackExchange, Wikipedia, Github. 
Following existing work~\cite{DBLP:journals/corr/abs-2308-06259,DBLP:journals/corr/abs-2304-08460}, we also sample documents from Open Assistant\footnote{\url{https://huggingface.co/datasets/OpenAssistant/oasst1}} and WikiHow\footnote{\url{https://huggingface.co/datasets/wikihow}} to introduce some structured examples.
We randomly choose several consecutive paragraphs from each original text in the corpus to serve as our document.
To ensure that each document contains enough information to generate at least one instruction-response pair, we only keep the documents that contain a range of 500 to 1000 tokens.

\subsection{Instruction Wrapper Building}
To empower a general LLM with the capability of instruction wrapping, we need to construct sufficient training examples mapping the document $\mathcal{D}$ to the instruction-response pair $(\mathcal{X}, \mathcal{Y})$. 
Inspired by \citep{DBLP:journals/corr/abs-2308-06259}, we leave this job to the well-aligned GPT-4~\citep{Chatgpt} to minimize the cost of human annotations.
We hypothesize that an ideal meta-training set $\Omega$ should fulfill two essential requirements: \textit{alignment} and \textit{diversity}. 
\textit{Alignment} guarantees that $\Omega$ encompasses a wide range of real human-written documents, enabling the wrapper to comprehend different domains and writing styles. 
\textit{Diversity}, on the other hand, ensures that $\Omega$ contains a variety of instructions, enabling the wrapper to generate diverse instructions effectively after training.
Therefore, we collect examples of $\Omega$ from the following two views.

\noindent \textbf{Alignment-view Examples.} 
In this section, the examples are constructed by utilizing GPT-4 to directly generate instruction-response pairs for real human-written documents.
To accomplish this goal, we harness the power of \textit{in-context learning} (ICL). 
In particular, for each domain, 30 examples are first manually constructed as the seeds. 
Then, for each $\mathcal{D}$, the prompt fed to GPT-4 is denoted by $(\mathcal{X}^*, \mathcal{D}_1, \mathcal{P}_1, ..., \mathcal{D}_k, \mathcal{P}_k, \mathcal{D})$, where $\mathcal{X}^*$ is definition of mapping $\mathcal{D}_j$ to the instruction-response pair $\mathcal{P}_j = (\mathcal{X}_j, \mathcal{Y}_j)$. 
See Appendix~\ref{app:full_prompt_1} for the full prompt. 
The resulting examples are denoted by $\Omega_a$.

\noindent \textbf{Diversity-view Examples.}
Intuitively, it is difficult to generate diverse instructions using just a few dozen manual seeds.
Therefore, we start from the publicly released instructions, such as \textsc{Alpaca}, and then inversely fuse their provided instructions and responses to create pseudo-documents. 
Specifically, for each instruction-response pair $(\mathcal{X}, \mathcal{Y})$ sampled in \textsc{Alpaca}, we write the prompt to employ GPT-4 to integrate $\mathcal{X}$ and $\mathcal{Y}$ into a new document $\tilde{\mathcal{D}}$. 
We enable $\tilde{\mathcal{D}}$ to encompass all the content from $\mathcal{X}$ and $\mathcal{Y}$, but we intentionally blur their boundaries. 
This allows for the addition of new content as needed, while ensuring a smooth and coherent flow of information.
These pseudo-documents $\tilde{\mathcal{D}}$ and their corresponding $(\mathcal{X}, \mathcal{Y})$ constitute the remaining training examples, denoted by $\Omega_d$.
Appendix~\ref{app:full_prompt_2} gives the detail prompt.

\noindent \textbf{Wrapper Training.}
we select \textsc{Llama2}~\citep{DBLP:journals/corr/abs-2307-09288}, an advanced LLM publicly available as our instruction wrapper $\mathcal{M}$ and perform \textit{supervised fine-tuning} (SFT) on $\mathcal{M}$ using the constructed meta-training set $\Omega = \Omega_a \cup \Omega_d$. 
For each document $D$ and its instruction-response pair $\mathcal{P} = (\mathcal{X}, \mathcal{Y})$ of $\Omega$, we add a unified instruction $\mathcal{U} = $ "\texttt{\footnotesize Convert the given text into a task. Input is a text and Response contains two fields: \#instruction\# and \#output\#.}". 
Then, the training loss is calculated by a log-likelihood, 
\begin{equation}
\label{eq:loss}
    \mathcal{L}(\mathcal{U}, \mathcal{D}, \mathcal{P}) =
    -{\sum_{j=1}^{|\mathcal{P}|} \log P(t_{j}|\mathcal{U}, \mathcal{D}, t_{< j})},
\end{equation}
where $t_{j}$ is the $j$-th token of $\mathcal{T}$.
It is crucial to emphasize that although our meta-training set $\Omega$ may include hallucinations, we hypothesize that this does not affect the learning of the wrapper $\mathcal{M}$. 
This is because our primary objective for $\mathcal{M}$  is to learn the stylistic transformation from documents to instruction-response pairs with semantic consistency.
During the inference phase, we exclusively utilize real human-written documents without pseudo-documents, which naturally reduces the occurrence of hallucinations.

\subsection{Data Generation via Instruction Wrapper}
In this stage, we use the trained $\mathcal{M}$ to generate instruction-response pairs for 20,000 human-written documents, which have been sampled using the method described in Section~\ref{sec:corpus}.
To avoid the hallucination that the wrapper generates too much content unrelated to the original text, we propose a post-processing strategy for each generated task $\mathcal{T}_i$. 
Concretely, we devise a score $\sigma(\mathcal{T}_i) = \min (\tilde{\sigma}(\mathcal{P}_i, \mathcal{X}_i), \tilde{\sigma}(\mathcal{P}_i, \mathcal{Y}_i))$ to measure the similarity between the text and the instruction-response pair, where $\tilde{\sigma}(\mathcal{P}_i, s) = |t(\mathcal{D}_i) \& t(s)| / |t(s)|$ and $t(s)$ denotes the token set of text $s$. All examples $(\mathcal{P}_i, \mathcal{T}_i)$ will be removed where $\sigma(\mathcal{T}_i) < \theta$.

\begin{table} 
	\begin{center}
	{\caption{Statistics of alignment-view examples $\Omega_a$, diversity-view examples $\Omega_d$, the meta-training set $\Omega$ and \textsc{DoG-Instruct}. Here $x \pm y$ denotes the average $x$ and standard deviation $y$. }\label{tab:statistics}}
 \scalebox{0.83}{
		\begin{tabular}{lccc}
			\toprule
            & \textbf{Example \#} & \textbf{$\mathcal{X}$ Token \#} & \textbf{$\mathcal{Y}$ Token \#} \\
		    \midrule
              $\Omega_a$ & $306$ & $16 \pm 13$ & $134 \pm 126$ \\
              $\Omega_d$ & $2998$ & $43 \pm 35$ & $140 \pm 76$ \\
            $\Omega$ & $3371$ & $41 \pm 34$ & $139 \pm 81$ \\
            \textsc{DoG-Instruct} & $12426$ & $32\pm79$ & $310 \pm 152$ \\
			\bottomrule
		\end{tabular}
  }
	\end{center}
\end{table}

\begin{table*} 
	\begin{center}
	{\caption{Statistics of different domains in \textsc{DoG-Instruct}. Instruction, input, and output lengths are given as the number of tokens. BS denotes the Bert-Score\citep{zhang2019bertscore}. OASST is short for Open Assistant.}\label{tab:relevance}}
 \scalebox{0.9}{
		\begin{tabular}{lccccccc}
            \toprule
             & Wikipedia & FreeLaw & ArXiv & StackExchange & Github & OASST & WikiHow \\
            \cmidrule(lr){1-1} \cmidrule(lr){2-8}
            \textbf{\# of Examples} & $5371$ & $427$ & $450$ & $475$ & $690$ & $946$ & $4060$ \\
            \textbf{Length of} $\mathcal{X}$ & $15 \pm 34$ & $201 \pm 207$ & $119 \pm 187$ & $101 \pm 104$ & $54 \pm 110$ & $34 \pm 51$ & $11 \pm 25$ \\
            \textbf{Length of} $\mathcal{Y}$ & $347 \pm 123$ & $476 \pm 123$ &  $577 \pm 205$ & $328 \pm 121$ & $328 \pm 132$ & $326 \pm 121$ & $326 \pm 104$ \\
            \cmidrule(lr){1-1} \cmidrule(lr){2-8}
            $\tilde{\sigma}(\mathcal{D}, \mathcal{Y})$ & $0.981$ & $0.949$ & $0.942$ & $0.976$ & $0.978$ & $0.957$ & $0.957$ \\
            BS$(\mathcal{D}, \mathcal{Y})$ & $0.981$ & $0.963$ & $0.942$ & $0.911$ & $0.967$ & $0.946$ & $0.930$ \\
			\bottomrule
		\end{tabular}
  }
	\end{center}
\end{table*}

\begin{table*} 
	\begin{center}
	{\caption{Performance of the methods on the AlpacaEval benchmark (win rate over text-davinci-003 evaluated by GPT-4). The \textbf{Text-Grounded} field indicates whether the instruction generation is based on human-written text. The \textbf{Avg. Length} denotes the average token number of the model responses. Our \textsc{DoG-Instruct} achieves the highest win rate (53.1\%) with the least training examples (12.4K).}\label{tab:alpaca_eval}}
	\scalebox{0.85}{
		\begin{tabular}{llccccc}
			\toprule
            \textbf{Data Generator} &\textbf{Dataset} &\textbf{Text-Grounded} &\textbf{\# of Examples} & \textbf{Win Rate (\%)} & \textbf{Avg. Length} \\
            \cmidrule(lr){1-1} \cmidrule(lr){2-2} \cmidrule(lr){3-3} \cmidrule(lr){4-4} \cmidrule(lr){5-6}
            \multirow{1}[1]{*}{\makecell[c]{text-davinci-003}}
            &\textsc{LongForm} &\checkmark &$23.7$K & $11.7$ & $268$  \\
			\cmidrule(lr){1-1} \cmidrule(lr){2-2} \cmidrule(lr){3-3} \cmidrule(lr){4-4} \cmidrule(lr){5-6}
            \multirow{5}[1]{*}{\makecell[c]{GPT-3.5-Turbo}}
            &\textsc{Self-Instruct} &$\times$ &$82$K &$14.2$ & $284$ \\
            &\textsc{Alpaca} &$\times$ &$52$K & $15.3$ & $271$ \\
            &\textsc{Dynosaur} &$\times$ &$800$K & $2.9$ & $142$  \\
            &\textsc{Evol-Instruct} &$\times$ & $70$K & $48.3$ & $669$ \\ 
            \cmidrule(lr){1-1} \cmidrule(lr){2-2} \cmidrule(lr){3-3} \cmidrule(lr){4-4} \cmidrule(lr){5-6}
            \multirow{1}[1]{*}{\makecell[c]{\textsc{GPT-4}}}
            &\textsc{Alpaca-GPT-4} &$\times$ &$52$K & $44.5$ & $653$ \\
            \cmidrule(lr){1-1} \cmidrule(lr){2-2} \cmidrule(lr){3-3} \cmidrule(lr){4-4} \cmidrule(lr){5-6}
            \multirow{2}[1]{*}{\makecell[c]{\textsc{Llama2-7B}}}
            &$\textsc{Humpback}^{\dag}$  &\checkmark &$18$K & $41.0$ & $755$ \\
            \cmidrule(lr){2-2} \cmidrule(lr){3-3} \cmidrule(lr){4-4} \cmidrule(lr){5-6}
            &\textsc{DoG-Instruct} &\checkmark &$12.4$K & $\mathbf{53.1}$ & $\mathbf{1149}$ \\
			\bottomrule
		\end{tabular}
		}
	\end{center}
\end{table*}

\begin{figure}
	\centering 
		\label{level.sub.2}
		\includegraphics[width=\linewidth]{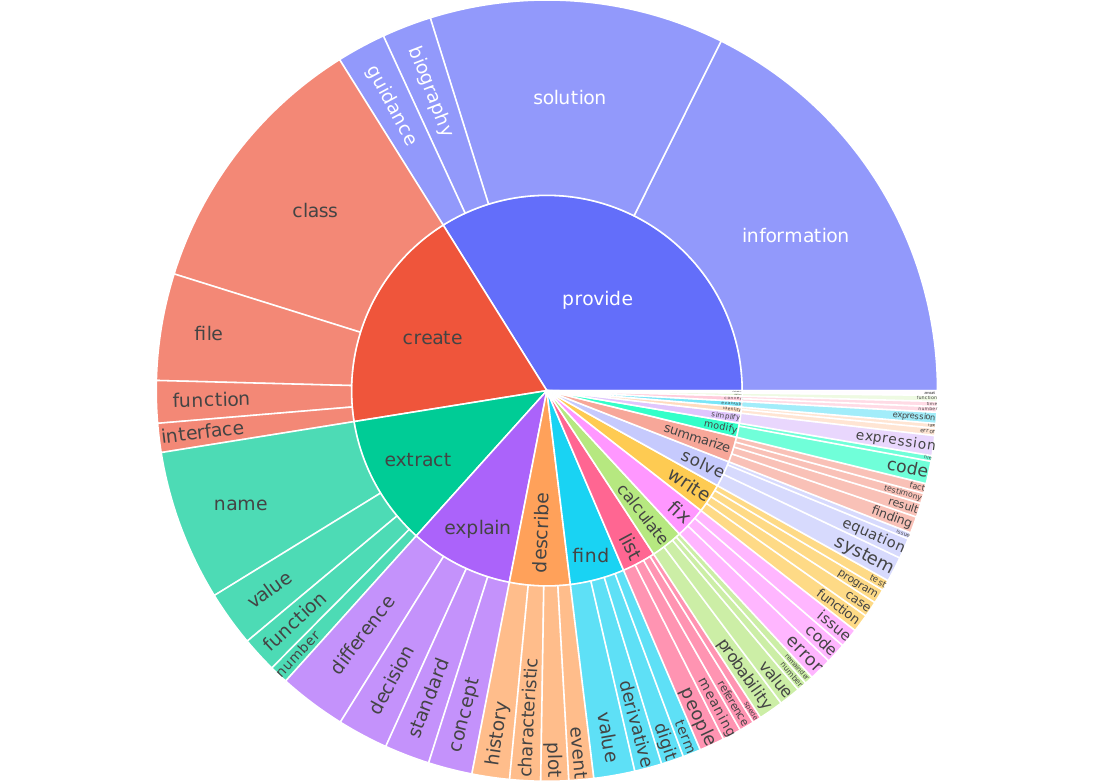}
  \caption{Instruction diversity of \textsc{DoG-Instruct} data. The inner circle shows common root verbs with the corresponding common noun objects in the outer circle.}\label{fig:diversity}
\end{figure}

\section{DoG-Instruct Statistics}
\noindent \textbf{Data Statistics.} 
Table \ref{tab:statistics} shows the statistics of alignment-view examples $\Omega_a$, diversity-view examples $\Omega_d$, the meta-training set $\Omega$, and \textsc{DoG-Instruct} dataset. 
Theoretically, as long as there is a constant stream of text, our method has no upper limit on the amount of data.
However, through experimentation, we discovered that competitive results can be achieved by using a mere 12k of our \textsc{DoG-Instruct}. 
\textsc{DoG-Instruct} tends to have longer responses compared to the examples in $\Omega$.
In addition, \textsc{DoG-Instruct} have larger standard deviations regarding the response field than $\Omega$.
The top of Table \ref{tab:relevance} presents the statistical data for different domains in \textsc{DoG-Instruct}. 

\noindent \textbf{Diversity of Instructions.} 
We performed a diversity analysis on \textsc{DoG-Instruct} using the method described by \cite{DBLP:conf/acl/WangKMLSKH23}. 
Figure \ref{fig:diversity} illustrates the distribution of the verb-noun structure of instructions, showcasing the diverse range. 

\noindent \textbf{Relevance to Raw Documents.}
Additionally, we computed the relevance of the responses to the raw documents. 
The average relevance scores are displayed at the bottom of Table \ref{tab:relevance}, with $\tilde{\sigma}$ representing the measure of literal relevance utilized in our post-processing, and BS denoting Bert-Score~\citep{zhang2019bertscore} for evaluating the semantic relevance.
Both in terms of literal score and semantic score, the responses exhibit a high level of relevance to the raw documents. However, they are not 100\% aligned due to the appropriate rewriting carried out by our instruction wrapper.

\section{Experiments}
\subsection{Experimental Setup}
\noindent \textbf{Compared Datasets.} We compared our \textsc{DoG-Instruct} with several typical instruction-tuning datasets: \textsc{Self-Instruct}~\citep{DBLP:conf/acl/WangKMLSKH23}, and \textsc{Alpaca}~\citep{alpaca} are automatically generated by LLMs including GPT-3.5-Turbo and text-davinci-003. \textsc{Dynosaur}~\citep{DBLP:journals/corr/abs-2305-14327} repackages huggingface's existing NLP dataset and regenerates instructions for it using ChatGPT. 
\textsc{LongForm}~\citep{DBLP:journals/corr/abs-2304-08460} and \textsc{Humpback}~\citep{DBLP:journals/corr/abs-2308-06259} are most similar to our work in that they generate tasks by performing the instruction back-translation. 
Unlike these methods, \textsc{DoG-Instruct} wraps the documents and carefully selects the valuable parts to compose a comprehensive instruction-response pair. 
Since \textsc{Humpback} hasn't been released yet, we got an unofficial version~\footnote{\url{https://huggingface.co/datasets/Spico/Humback}} from HuggingFace, denoted by ${\dag}$.

\begin{table*} 
	\begin{center}
	{\caption{Rouge-L (R), Meteor (M), and Bert-Score (B) of different methods on the test sets of three benchmarks. All methods follow zero-shot settings.}\label{tab:benchmark}}
	\scalebox{0.85}{
		\begin{tabular}{llcccccc}
			\toprule
            \multicolumn{1}{l}{\multirow{2}[1]{*}{\makecell[c]{\textbf{Data Generator}}}}
			&\multicolumn{1}{l}{\multirow{2}[1]{*}{\textbf{Dataset}}}
            &\multicolumn{1}{l}{\multirow{2}[1]{*}{\makecell[c]{\textbf{\# of Examples}}}}
			&\multicolumn{2}{c}{\textbf{ELI5}}&\multicolumn{2}{c}{\textbf{LF-Test}}&\multicolumn{1}{c}{\textbf{Super-NI}} \\
		    \cmidrule(lr){4-5} \cmidrule(lr){6-7}  \cmidrule(lr){8-8}
			
			&&& R(\%) & M(\%)  & R(\%) & M(\%) & B(\%)  \\
            \cmidrule(lr){1-1} \cmidrule(lr){2-2} \cmidrule(lr){3-3} \cmidrule(lr){4-5} \cmidrule(lr){6-7} \cmidrule(lr){8-8}
            \cmidrule(lr){1-1} \cmidrule(lr){2-2} \cmidrule(lr){3-3} \cmidrule(lr){4-5} \cmidrule(lr){6-7} \cmidrule(lr){8-8}
            \multirow{1}[1]{*}{\makecell[c]{text-davinci-003}}
            &\textsc{LongForm} &$23.7$K & $7.5$ & $5.4$ & $24.9$ & $18.1$ & $81.8$ \\
            \cmidrule(lr){1-1} \cmidrule(lr){2-2} \cmidrule(lr){3-3} \cmidrule(lr){4-5} \cmidrule(lr){6-7} \cmidrule(lr){8-8}
            \multirow{5}[1]{*}{\makecell[c]{GPT-3.5-Turbo}}
            &\textsc{Self-Instruct} &$82$K & $9.8$ & $8.2$ & $22.4$ & $16.5$ & $83.0$ \\
            &\textsc{Alpaca} &$52$K & $10.1$ & $8.8$ & $23.1$ & $17.3$ & $82.9$  \\
            &\textsc{Dynosaur} &$800$K & $3.1$ & $1.5$  &$15.6$ & $11.0$ & $86.0$ \\
            &\textsc{Evol-Instruct} & $70$K & $18.9$ & $18.4$  & $25.2$ & $21.8$ & $85.6$  \\ 
            \cmidrule(lr){1-1} \cmidrule(lr){2-2} \cmidrule(lr){3-3} \cmidrule(lr){4-5} \cmidrule(lr){6-7} \cmidrule(lr){8-8}
            \multirow{1}[1]{*}{\makecell[c]{\textsc{GPT-4}}}
            &\textsc{Alpaca}-GPT-4 &$52$K & $11.1$ & $13.3$  &$25.1$ & $22.4$ & $85.8$   \\
            \cmidrule(lr){1-1} \cmidrule(lr){2-2} \cmidrule(lr){3-3} \cmidrule(lr){4-5} \cmidrule(lr){6-7} \cmidrule(lr){8-8}
            \multirow{2}[1]{*}{\makecell[c]{\textsc{Llama2-7B}}}
            &$\textsc{Humpback}^{\dag}$ &$18$K & $9.3$ & $6.1$  & $25.0$ & $22.2$ & $83.7$ \\
            \cmidrule(lr){2-2} \cmidrule(lr){3-3} \cmidrule(lr){4-5} \cmidrule(lr){6-7} \cmidrule(lr){8-8}
            &\textsc{DoG-Instruct} &$12.4$K & $\mathbf{19.0}$ & $\mathbf{19.7}$ & $\mathbf{25.9}$ & $\mathbf{23.6}$ & $\mathbf{86.1}$  \\
			\bottomrule
		\end{tabular}
		}
	\end{center}
\end{table*}

\noindent \textbf{Implementation Details.}
All our experiments ran on 8 Tesla V100 GPUs with FP16. We trained $\mathcal{M}$ using \textsc{LoRA}~\citep{DBLP:conf/iclr/HuSWALWWC22}. The hyper-parameters were set as follows: 
(1) The batch size was set to $128$. 
(2) The learning rate was set to $1 \times 10^{-4}$. 
(3) The epoch number was $7$. 
(4) The cutoff token number was set to $2048$. 
(5) The temperature and beam size were $0$ and $4$, respectively.
(6) The \textsc{LoRA} target modules consisted of $[\text{q}_\text{proj},\text{k}_\text{proj},\text{v}_\text{proj},\text{o}_\text{proj},\text{up}_\text{proj},\text{down}_\text{proj},\text{gate}_\text{proj},\\ \text{embed}_\text{tokens},\text{lm}_\text{head}]$.

\begin{table*} 
	\begin{center}
	{\caption{Experimental results of ablation studies for all benchmarks used.}\label{tab:ablation_test}}
 \scalebox{0.9}{
		\begin{tabular}{llcccc}
            \toprule
            \multicolumn{1}{l}{\multirow{3}[1]{*}{Stage}}
            &\multicolumn{1}{l}{\multirow{2}[1]{*}{Setting}}&\multicolumn{1}{c}{\textbf{AlpacaEval}}
			&\multicolumn{1}{c}{\textbf{ELI5}}&\multicolumn{1}{c}{\textbf{LF-Test}}&\multicolumn{1}{c}{\textbf{Super-NI}} \\
            \cmidrule(lr){3-3} \cmidrule(lr){4-4} \cmidrule(lr){5-5} \cmidrule(lr){6-6}
            && Win Rate(\%) & M(\%)  & M(\%) & B(\%)  \\
            \cmidrule(lr){2-2} \cmidrule(lr){3-3} \cmidrule(lr){4-4} \cmidrule(lr){5-5} \cmidrule(lr){6-6}
            &\textsc{DoG-Instruct} & $\mathbf{53.1}$ & $\mathbf{19.7}$ & $\mathbf{23.6}$ & $\mathbf{86.1}$ \\
            \cmidrule(lr){1-1} \cmidrule(lr){2-2} \cmidrule(lr){3-3} \cmidrule(lr){4-4} \cmidrule(lr){5-5} \cmidrule(lr){6-6}
            \multirow{3}[1]{*}{\makecell[c]{ Training}} 
	       &w/o alignment-view & $46.7$ & $18.5$ & $20.2$ & $85.2$ \\   
             &w/o diversity-view & $12.5$ & $12.1$ & $15.7$ & $83.8$ \\   
             &w instruction back-translation & $5.9$ & $9.3$ & $16.6$ & $81.7$ \\
       \cmidrule(lr){1-1} \cmidrule(lr){2-2} \cmidrule(lr){3-3} \cmidrule(lr){4-4} \cmidrule(lr){5-5}	\cmidrule(lr){6-6}
       \multirow{1}[1]{*}{\makecell[c]{Generation}} &w/o post-processing & $32.0$ & $17.2$ & $23.3$ & $85.9$ \\
             \bottomrule
		\end{tabular}
  }
	\end{center}
\end{table*}

\subsection{Automatic Evaluation}
To begin with, we conducted an automatic evaluation on multiple benchmarks. 
For each dataset being compared, We independently fine-tuned an identical baseline LLM using its respective training examples and evaluated its performance in accurately following the instructions.

\noindent \textbf{Baseline LLM.} 
We select \textsc{Llama2-7B}~\citep{DBLP:journals/corr/abs-2307-09288} + \textsc{LoRA}~\citep{DBLP:conf/iclr/HuSWALWWC22} as the baseline LLM. For ease of presentation, we refer to the baseline LLM trained on dataset $x$ as $x$-model.

\noindent \textbf{Benchmarks.} 
We first used the GPT-4 evaluation from AlpacaEval~\cite{li2023alpacaeval} to evaluate response quality on 805 instructions from the Alpaca Leaderboard. 
AlpacaEval compares the pairwise win rate against the reference model text-davinci-003.
In addition, we conducted evaluations on three other NLG benchmarks: Long-Form Question Answering (ELI5)~\citep{DBLP:conf/acl/FanJPGWA19}, LongForm test set (LF-Test)~\citep{DBLP:journals/corr/abs-2304-08460}, and Super-NaturalInstructions (Super-NI)~\citep{DBLP:conf/emnlp/WangMAKMNADASPK22}. 
None of the methods incorporate data from these benchmarks. i.e. zero-shot setting.

\noindent \textbf{Automatic Evaluation Metrics.} 
For AlpacaEval, we ran its scripts directly, using GPT-4 for evaluation.
For ELI5 and LF-Test, we followed \citep{DBLP:journals/corr/abs-2304-08460,DBLP:journals/corr/abs-2305-14327} to calculate the Rouge-L~\citep{lin2004rouge} and Meteor~\citep{DBLP:conf/acl/BanerjeeL05} scores. 
These scores are computed by comparing the model outputs with the provided references in the respective datasets. 
For Super-NI, we utilize Bert-Score~\citep{zhang2019bertscore} for evaluation instead of other long-text metrics like Rouge. 
This choice is made due to the typically short nature of the outputs in this dataset.

\subsubsection{AlpacaEval Results.} 
The win rate and average length of model responses for different methods on AlpacaEval are presented in Table~\ref{tab:alpaca_eval}. 
It is worth highlighting that despite utilizing the least amount of data, we achieved the best performance while maintaining the same baseline LLM premise at the 7B model scale.
The \textsc{Dynosaur}-model demonstrates the lowest performance, potentially due to its output being excessively standardized and concise rather than a detailed reply.
By surpassing all non-text-based methods, we demonstrate the effectiveness of human-written text in mitigating LLM hallucinations.
In comparison to the text-grounded method Humpback, we achieved a substantial improvement by adapting our command wrapper to the AI response style, resulting in significant advancements.

\begin{figure*}
\centering
    \begin{tikzpicture}[scale=0.6]
        \begin{axis}[
        ybar,
        bar width=10pt,
        enlargelimits=0.4,
        symbolic x coords={\textsc{Evol}, \textsc{Humpback}, \textsc{DoG}},
    	xtick=data,
        ymajorgrids=true, 
        grid style=dashed,
        tick label style={font=\large},
        label style={font=\large}]
    	
        \addplot[draw=blue, fill=blue!30!white]
    	coordinates {(\textsc{Evol}, 58.7) (\textsc{Humpback}, 33.9) (\textsc{DoG}, 66.6) };

        \addplot[draw=cyan, fill=cyan!30!white]
    	coordinates {(\textsc{Evol}, 5.4) (\textsc{Humpback}, 7.5) (\textsc{DoG}, 14.6) };

       \addplot[draw=purple, fill=purple!30!white]
    	coordinates {(\textsc{Evol}, 35.9) (\textsc{Humpback}, 58.6) (\textsc{DoG}, 18.8) };
        \end{axis}
    \end{tikzpicture}
    \begin{tikzpicture}[scale=0.6]
        \begin{axis}[
        ybar,
        bar width=10pt,
        enlargelimits=0.4,
        symbolic x coords={\textsc{Evol}, \textsc{Humpback}, \textsc{DoG}},
    	xtick=data,
        ymajorgrids=true, 
        grid style=dashed,
        tick label style={font=\large},
        label style={font=\large}]
    	
        \addplot[draw=blue, fill=blue!30!white]
    	coordinates {(\textsc{Evol}, 52.3) (\textsc{Humpback}, 50.31) (\textsc{DoG}, 65.31) };

        \addplot[draw=cyan, fill=cyan!30!white]
    	coordinates {(\textsc{Evol}, 18.8) (\textsc{Humpback}, 14.33) (\textsc{DoG}, 12.8) };

       \addplot[draw=purple, fill=purple!30!white]
    	coordinates {(\textsc{Evol}, 28.9) (\textsc{Humpback}, 35.36) (\textsc{DoG}, 21.89) };
        \end{axis}
    \end{tikzpicture}
    \begin{tikzpicture}[scale=0.6]
        \begin{axis}[
        ybar,
        bar width=10pt,
        enlargelimits=0.4,
        legend style={font=\Large, at={(1.15,0.95)}},
        symbolic x coords={\textsc{Evol}, \textsc{Humpback}, \textsc{DoG}},
    	xtick=data,
        ymajorgrids=true, 
        grid style=dashed,
        tick label style={font=\large},
        label style={font=\large}]
    	
        \addplot[draw=blue, fill=blue!30!white]
    	coordinates {(\textsc{Evol}, 45.2) (\textsc{Humpback}, 22.7) (\textsc{DoG}, 48.7) };
    	\addlegendentry{Win Rate}

        \addplot[draw=cyan, fill=cyan!30!white]
    	coordinates {(\textsc{Evol}, 7.1) (\textsc{Humpback}, 11.9) (\textsc{DoG}, 7.6) };
    	\addlegendentry{Tie Rate}

       \addplot[draw=purple, fill=purple!30!white]
    	coordinates {(\textsc{Evol}, 47.7) (\textsc{Humpback}, 65.4) (\textsc{DoG}, 43.7) };
    	\addlegendentry{Lose Rate}
        \end{axis}
    \end{tikzpicture}

\caption{GPT-4 automatic evaluation results on subsets of Eli5 (left), LF-Test (middle), Super-NI (right). To account for the cost of GPT-4, each subset contains 200 examples that randomly sampled from the original test sets. The win/tie/lose rates are computed by comparing the model responses with the given reference responses.}
\label{fig:gpt4_eval}
\end{figure*}
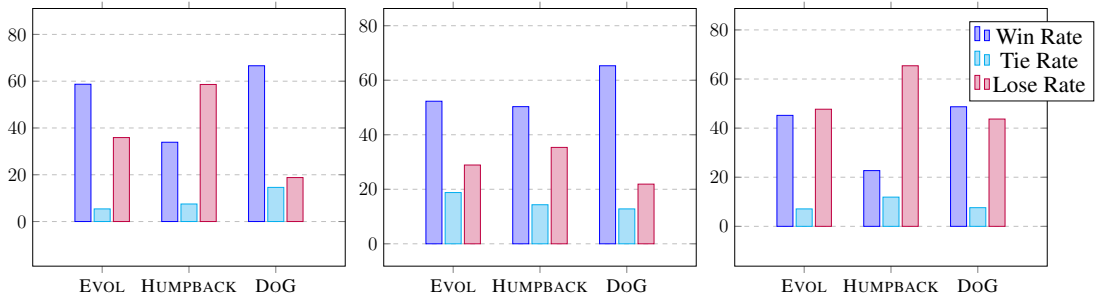

\subsubsection{ELI5, LF-Test and Super-NI Results.} 
Table~\ref{tab:benchmark} shows the Rouge-L (R), Meteor (M), and Bert-Score (B) of different models on ELI5, LF-Test, and Super-NI. 
Our method outperforms all the compared methods across all three benchmarks in terms of Rouge-L, Meteor, and Bert-Score, achieving superior performance in all evaluation metrics. 
This observation showcases that our dataset enables better alignment between LLM outputs and human annotations, indicating the efficacy of our method in improving the performance of LLM models.

\noindent \textbf{GPT-4 Evaluation.}
To mitigate any bias introduced by conventional metrics such as Rouge, we employed GPT-4 for evaluation on ELI5, LF-Test, and Super-NI benchmarks. We calculated the win/tie/lose rates by comparing the model responses with the reference responses provided by the benchmarks.
The results are shown in Figure~\ref{fig:gpt4_eval}. Our \textsc{DoG-instruct}-model consistently achieves the highest win rate across all benchmarks.

\subsubsection{Ablation Study.} 
We compared the LLM performance using different settings to construct \textsc{DoG-instruct}.

\begin{itemize}[leftmargin=1em]
    \item \textbf{w/o alignment-view}: we reconstructed the meta-training set $\Omega$ without any examples constructed by real human-written texts;
    \item \textbf{w/o diversity-view}: we reconstructed $\Omega$ without any examples fused by the instructions and responses from \textsc{Alpaca};
    \item \textbf{w instruction back-translation}: we replaced our instruction wrapping with instruction back-translation to reconstruct \textsc{DoG-Instruct} while keeping the input documents unchanged.
    \item \textbf{w/o post-processing}: we removed post-processing when generating \textsc{DoG-Instruct}.
\end{itemize}

\begin{table} 
	\begin{center}
	{\caption{Human evaluation on dataset qualification. 
 For each dataset, we randomly sampled 50 examples. Here $\downarrow$ means the smaller the value, the better.}\label{tab:human_eval}}
    \scalebox{0.88}{
		\begin{tabular}{lccc}
			\toprule
            Dataset &  V (\%) & H (\%) $\downarrow$ & F (\%) \\
		    \midrule
            \textsc{Alpaca}-GPT-4 &$94$ & $22$ & $92$ \\
            \textsc{Evol-Instruct} & $94$ & $18$ & $94$ \\
            \textsc{LongForm} & $76$ &  $14$ & $84$ \\
            $\textsc{Humpback}^{\dag}$ & $48$ & $12$ & $62$ \\
            \midrule
            $\Omega$ & $92$ & $20$ & $94$ \\
            \textsc{DoG-Instruct} & $\textbf{96}$ & $\textbf{12}$ & $\textbf{96}$ \\
			\bottomrule
		\end{tabular}
    }
	\end{center}
\end{table}

Table \ref{tab:ablation_test} shows the experimental results. Our \textsc{DoG-Instruct} equipped with all components performs best in terms of all metrics. 
Dramatic performance degradation demonstrates that the adaptation of the PLM to the task format is critical to the effectiveness of prompt tuning.

\subsection{Human Evaluation}
\label{sec:human_eval}
While the automatic evaluation in the previous section provided an overall assessment of model performance, we now aim to specifically evaluate the effectiveness of our \textsc{DoG-Instruct} in reducing hallucinations and aligning model responses with human-like outputs.
\subsubsection{Data Quality}
We randomly select 50 examples from each dataset and manually evaluate their quality. 
Since the generated tasks may involve knowledge from several different domains, we require that the annotator needs to retrieve the corresponding evidence using the search engines and compare them one by one. 
The entire process of manual evaluation took approximately 8 man-hours.

\noindent \textbf{Human Evaluation Metrics.}
a) \textit{validation} (V) indicates the percentage of the example whose response follows the instruction.
b) \textit{hallucination} (H) measures the percentage of the example whose response contains factual errors. 
c) \textit{fluency} (F) indicates the percentage of the example that has instructions and responses that are smooth and fluent.

\begin{figure}
\centering
    \begin{tikzpicture}[scale=0.7]
        \begin{axis}[
        ybar,
        bar width=10pt,
        enlargelimits=0.2,
        legend style={font=\Large, at={(1.4,0.95)}},
        symbolic x coords={GPT-4, \textsc{LongForm}, \textsc{Humpback}},
    	xtick=data,
        ymajorgrids=true, 
        grid style=dashed,
        tick label style={font=\large},
        label style={font=\Large}]
    	
        \addplot[draw=brown, fill=brown!30!white]
    	coordinates {(GPT-4, 32.5) (\textsc{LongForm}, 24.2) (\textsc{Humpback}, 67.0) };
    	\addlegendentry{$\mathcal{M}$ Wins}

        \addplot[draw=yellow, fill=yellow!30!white]
    	coordinates {(GPT-4, 35.5) (\textsc{LongForm}, 63.1) (\textsc{Humpback}, 28.2) };
    	\addlegendentry{Tie}

       \addplot[draw=orange, fill=orange!30!white]
    	coordinates {(GPT-4, 32) (\textsc{LongForm}, 12.7) (\textsc{Humpback}, 4.8) };
    	\addlegendentry{$\mathcal{M}$ Loses}
        \end{axis}
    \end{tikzpicture}

\caption{Human evaluation comparing \textsc{DoG-Instruct} with various text-grounded methods. The evaluation was carried out using the same set of human-written documents as input for all methods.}\label{fig:win_lose}
\end{figure}
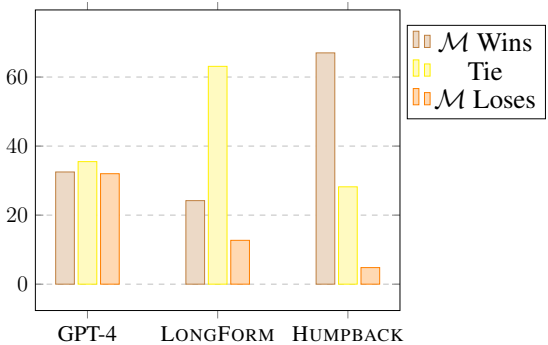

\noindent \textbf{Results.}  The results are shown in Figure \ref{tab:human_eval}. 
Both \textsc{Alpaca}-GPT-4 and \textsc{Evol-Instruct} demonstrate higher levels of hallucination due to their complete reliance on using LLMs to generate instruction-response pairs from scratch.
By generating tasks from human-written documents, both \textsc{LongForm} and \textsc{Humpback} effectively mitigate the hallucinations. 
Nevertheless, the inclusion of noise in real text leads to lower fluency (F) compared to datasets that are fully generated by LLMs.
In contrast, our method combines the use of human-written text as factual support with style modification through LLMs, leading to superior performance across all three metrics.

\subsubsection{Text-Grounded Generation Capability} 
For the same document, we compared the quality of generated instruction-response pairs by employing two different methods: instruction wrapping and instruction back-translation.
Specifically, we randomly selected 100 documents from the corpus used in \textsc{LongForm} and \textsc{Humpback} to feed our instruction wrapper $\mathcal{M}$.
The results, depicted in Figure~\ref{fig:win_lose}, demonstrate that our wrapper yields instruction-response pairs of superior quality for the same document.
Furthermore, we randomly sampled approximately 100 documents from our and had GPT-4 to perform instruction wrapping.
Figure~\ref{fig:win_lose} illustrates that the instruction-response pairs generated by our $\mathcal{M}$ exhibit competitive quality to those produced by GPT-4.

\section{Related Work}
\textbf{Instruction Tuning} 
Humans possess the ability to effortlessly comprehend and execute tasks based on verbal instructions~\citep{DBLP:journals/corr/abs-2302-13971,Chatgpt,DBLP:journals/corr/abs-2307-09288}. Likewise, advancements in deep learning have enabled Language Models (LLMs)~\citep{DBLP:conf/nips/BrownMRSKDNSSAA20,Chatgpt,DBLP:journals/corr/abs-2204-02311,DBLP:journals/corr/abs-2302-13971} to acquire the capability to understand and follow instructions. Instruction tuning serves as a promising method, involving the fine-tuning of LLMs using training data and instructions from a collection of upstream tasks\citep{DBLP:conf/iclr/SanhWRBSACSRDBX22,DBLP:conf/acl/MishraKBH22,DBLP:conf/iclr/WeiBZGYLDDL22,DBLP:journals/corr/abs-2210-11416,DBLP:conf/icml/LongpreHVWCTZLZ23,DBLP:journals/corr/abs-2304-03277}. 

\noindent \textbf{Instruction-Tuning Data Collection} 
The collection of high-quality instruction-tuning data~\citep{DBLP:journals/corr/abs-2304-12244,DBLP:journals/corr/abs-2305-14327,DBLP:conf/acl/HonovichSLS23} is a pressing issue in enhancing the capability of instruction-following. Previous methods can be broadly categorized into three main groups. Firstly, methods like \textsc{Super-NI}~\citep{DBLP:conf/emnlp/WangMAKMNADASPK22} and \textsc{Dolly}~\citep{DatabricksBlog2023DollyV2} rely on hiring professionals to create instructions for diverse NLP tasks. 
Secondly, methods such as \textsc{Self-Instruct}~\citep{DBLP:conf/acl/WangKMLSKH23} and \textsc{Alpaca}~\citep{alpaca} advocate for the use of LLMs to automatically generate instruction-tuning data, thus eliminating the need for manual labor. 
Lastly, Dynosaur~\citep{DBLP:journals/corr/abs-2305-14327} employs LLMs to convert existing NLP datasets from Huggingface into instruction-tuning data at a reduced cost. 
The work most related to this paper is \cite{DBLP:journals/corr/abs-2304-08460,DBLP:journals/corr/abs-2308-06259}. 
It uses a human-written document as a natural response and leverages an LLM to generate the corresponding instruction based on the response. 
In contrast, our instruction wrapper selects the valuable parts of the documents for constructing appropriate responses. 

\section{Conclusion \& Limitation}
This paper introduces a new method called instruction wrapping, which enables the automatic collection of high-quality instruction-tuning data from human-written documents. 
Our trained instruction wrapper not only utilizes documents to mitigate response hallucinations but also modifies raw documents to align them with the standard response style. 
Through comprehensive evaluations, we demonstrate that our method achieves remarkable results on various widely used benchmarks while utilizing the fewest training examples.
The limitations of our method are that it cannot handle excessively long documents and can only generate a single task for a document.
In future work, we will explore generating more complicated instructions that involve multiple longer documents. 

\section*{Acknowledgement}
This work is partially supported by National Nature Science Foundation of China under No. U21A20488. We thank the Big Data Computing Center of Southeast University for providing the facility support on the numerical calculations in this paper.

\bibliography{anthology,custom}

\begin{thebibliography}{29}
\expandafter\ifx\csname natexlab\endcsname\relax\def\natexlab#1{#1}\fi

\bibitem[{Banerjee and Lavie(2005)}]{DBLP:conf/acl/BanerjeeL05}
Satanjeev Banerjee and Alon Lavie. 2005.
\newblock \href {https://aclanthology.org/W05-0909/} {{METEOR:} an automatic metric for {MT} evaluation with improved correlation with human judgments}.
\newblock In \emph{Proceedings of the Workshop on Intrinsic and Extrinsic Evaluation Measures for Machine Translation and/or Summarization@ACL 2005, Ann Arbor, Michigan, USA, June 29, 2005}, pages 65--72. Association for Computational Linguistics.

\bibitem[{Brown et~al.(2020)Brown, Mann, Ryder, Subbiah, Kaplan, Dhariwal, Neelakantan, Shyam, Sastry, Askell, Agarwal, Herbert{-}Voss, Krueger, Henighan, Child, Ramesh, Ziegler, Wu, Winter, Hesse, Chen, Sigler, Litwin, Gray, Chess, Clark, Berner, McCandlish, Radford, Sutskever, and Amodei}]{DBLP:conf/nips/BrownMRSKDNSSAA20}
Tom~B. Brown, Benjamin Mann, Nick Ryder, Melanie Subbiah, Jared Kaplan, Prafulla Dhariwal, Arvind Neelakantan, Pranav Shyam, Girish Sastry, Amanda Askell, Sandhini Agarwal, Ariel Herbert{-}Voss, Gretchen Krueger, Tom Henighan, Rewon Child, Aditya Ramesh, Daniel~M. Ziegler, Jeffrey Wu, Clemens Winter, Christopher Hesse, Mark Chen, Eric Sigler, Mateusz Litwin, Scott Gray, Benjamin Chess, Jack Clark, Christopher Berner, Sam McCandlish, Alec Radford, Ilya Sutskever, and Dario Amodei. 2020.
\newblock \href {https://proceedings.neurips.cc/paper/2020/hash/1457c0d6bfcb4967418bfb8ac142f64a-Abstract.html} {Language models are few-shot learners}.
\newblock In \emph{Advances in Neural Information Processing Systems 33: Annual Conference on Neural Information Processing Systems 2020, NeurIPS 2020, December 6-12, 2020, virtual}.

\bibitem[{Chowdhery et~al.(2022)Chowdhery, Narang, Devlin, Bosma, Mishra, Roberts, Barham, Chung, Sutton, Gehrmann, Schuh, Shi, Tsvyashchenko, Maynez, Rao, Barnes, Tay, Shazeer, Prabhakaran, Reif, Du, Hutchinson, Pope, Bradbury, Austin, Isard, Gur{-}Ari, Yin, Duke, Levskaya, Ghemawat, Dev, Michalewski, Garcia, Misra, Robinson, Fedus, Zhou, Ippolito, Luan, Lim, Zoph, Spiridonov, Sepassi, Dohan, Agrawal, Omernick, Dai, Pillai, Pellat, Lewkowycz, Moreira, Child, Polozov, Lee, Zhou, Wang, Saeta, Diaz, Firat, Catasta, Wei, Meier{-}Hellstern, Eck, Dean, Petrov, and Fiedel}]{DBLP:journals/corr/abs-2204-02311}
Aakanksha Chowdhery, Sharan Narang, Jacob Devlin, Maarten Bosma, Gaurav Mishra, Adam Roberts, Paul Barham, Hyung~Won Chung, Charles Sutton, Sebastian Gehrmann, Parker Schuh, Kensen Shi, Sasha Tsvyashchenko, Joshua Maynez, Abhishek Rao, Parker Barnes, Yi~Tay, Noam Shazeer, Vinodkumar Prabhakaran, Emily Reif, Nan Du, Ben Hutchinson, Reiner Pope, James Bradbury, Jacob Austin, Michael Isard, Guy Gur{-}Ari, Pengcheng Yin, Toju Duke, Anselm Levskaya, Sanjay Ghemawat, Sunipa Dev, Henryk Michalewski, Xavier Garcia, Vedant Misra, Kevin Robinson, Liam Fedus, Denny Zhou, Daphne Ippolito, David Luan, Hyeontaek Lim, Barret Zoph, Alexander Spiridonov, Ryan Sepassi, David Dohan, Shivani Agrawal, Mark Omernick, Andrew~M. Dai, Thanumalayan~Sankaranarayana Pillai, Marie Pellat, Aitor Lewkowycz, Erica Moreira, Rewon Child, Oleksandr Polozov, Katherine Lee, Zongwei Zhou, Xuezhi Wang, Brennan Saeta, Mark Diaz, Orhan Firat, Michele Catasta, Jason Wei, Kathy Meier{-}Hellstern, Douglas Eck, Jeff Dean, Slav Petrov, and Noah Fiedel.
  2022.
\newblock \href {https://doi.org/10.48550/arXiv.2204.02311} {Palm: Scaling language modeling with pathways}.
\newblock \emph{CoRR}, abs/2204.02311.

\bibitem[{Chung et~al.(2022)Chung, Hou, Longpre, Zoph, Tay, Fedus, Li, Wang, Dehghani, Brahma, Webson, Gu, Dai, Suzgun, Chen, Chowdhery, Narang, Mishra, Yu, Zhao, Huang, Dai, Yu, Petrov, Chi, Dean, Devlin, Roberts, Zhou, Le, and Wei}]{DBLP:journals/corr/abs-2210-11416}
Hyung~Won Chung, Le~Hou, Shayne Longpre, Barret Zoph, Yi~Tay, William Fedus, Eric Li, Xuezhi Wang, Mostafa Dehghani, Siddhartha Brahma, Albert Webson, Shixiang~Shane Gu, Zhuyun Dai, Mirac Suzgun, Xinyun Chen, Aakanksha Chowdhery, Sharan Narang, Gaurav Mishra, Adams Yu, Vincent~Y. Zhao, Yanping Huang, Andrew~M. Dai, Hongkun Yu, Slav Petrov, Ed~H. Chi, Jeff Dean, Jacob Devlin, Adam Roberts, Denny Zhou, Quoc~V. Le, and Jason Wei. 2022.
\newblock \href {https://doi.org/10.48550/arXiv.2210.11416} {Scaling instruction-finetuned language models}.
\newblock \emph{CoRR}, abs/2210.11416.

\bibitem[{Conover et~al.(2023)Conover, Hayes, Mathur, Xie, Wan, Shah, Ghodsi, Wendell, Zaharia, and Xin}]{DatabricksBlog2023DollyV2}
Mike Conover, Matt Hayes, Ankit Mathur, Jianwei Xie, Jun Wan, Sam Shah, Ali Ghodsi, Patrick Wendell, Matei Zaharia, and Reynold Xin. 2023.
\newblock \href {https://www.databricks.com/blog/2023/04/12/dolly-first-open-commercially-viable-instruction-tuned-llm} {Free dolly: Introducing the world's first truly open instruction-tuned llm}.

\bibitem[{Fan et~al.(2019)Fan, Jernite, Perez, Grangier, Weston, and Auli}]{DBLP:conf/acl/FanJPGWA19}
Angela Fan, Yacine Jernite, Ethan Perez, David Grangier, Jason Weston, and Michael Auli. 2019.
\newblock \href {https://doi.org/10.18653/v1/p19-1346} {{ELI5:} long form question answering}.
\newblock In \emph{Proceedings of the 57th Conference of the Association for Computational Linguistics, {ACL} 2019, Florence, Italy, July 28- August 2, 2019, Volume 1: Long Papers}, pages 3558--3567. Association for Computational Linguistics.

\bibitem[{Gao et~al.(2021)Gao, Biderman, Black, Golding, Hoppe, Foster, Phang, He, Thite, Nabeshima, Presser, and Leahy}]{DBLP:journals/corr/abs-2101-00027}
Leo Gao, Stella Biderman, Sid Black, Laurence Golding, Travis Hoppe, Charles Foster, Jason Phang, Horace He, Anish Thite, Noa Nabeshima, Shawn Presser, and Connor Leahy. 2021.
\newblock \href {http://arxiv.org/abs/2101.00027} {The pile: An 800gb dataset of diverse text for language modeling}.
\newblock \emph{CoRR}, abs/2101.00027.

\bibitem[{Honovich et~al.(2023)Honovich, Scialom, Levy, and Schick}]{DBLP:conf/acl/HonovichSLS23}
Or~Honovich, Thomas Scialom, Omer Levy, and Timo Schick. 2023.
\newblock \href {https://doi.org/10.18653/v1/2023.acl-long.806} {Unnatural instructions: Tuning language models with (almost) no human labor}.
\newblock In \emph{Proceedings of the 61st Annual Meeting of the Association for Computational Linguistics (Volume 1: Long Papers), {ACL} 2023, Toronto, Canada, July 9-14, 2023}, pages 14409--14428. Association for Computational Linguistics.

\bibitem[{Hu et~al.(2022)Hu, Shen, Wallis, Allen{-}Zhu, Li, Wang, Wang, and Chen}]{DBLP:conf/iclr/HuSWALWWC22}
Edward~J. Hu, Yelong Shen, Phillip Wallis, Zeyuan Allen{-}Zhu, Yuanzhi Li, Shean Wang, Lu~Wang, and Weizhu Chen. 2022.
\newblock \href {https://openreview.net/forum?id=nZeVKeeFYf9} {Lora: Low-rank adaptation of large language models}.
\newblock In \emph{The Tenth International Conference on Learning Representations, {ICLR} 2022, Virtual Event, April 25-29, 2022}. OpenReview.net.

\bibitem[{K{\"{o}}ksal et~al.(2023)K{\"{o}}ksal, Schick, Korhonen, and Sch{\"{u}}tze}]{DBLP:journals/corr/abs-2304-08460}
Abdullatif K{\"{o}}ksal, Timo Schick, Anna Korhonen, and Hinrich Sch{\"{u}}tze. 2023.
\newblock \href {https://doi.org/10.48550/arXiv.2304.08460} {Longform: Optimizing instruction tuning for long text generation with corpus extraction}.
\newblock \emph{CoRR}, abs/2304.08460.

\bibitem[{Li et~al.(2023{\natexlab{a}})Li, Yu, Zhou, Schick, Zettlemoyer, Levy, Weston, and Lewis}]{DBLP:journals/corr/abs-2308-06259}
Xian Li, Ping Yu, Chunting Zhou, Timo Schick, Luke Zettlemoyer, Omer Levy, Jason Weston, and Mike Lewis. 2023{\natexlab{a}}.
\newblock \href {https://doi.org/10.48550/arXiv.2308.06259} {Self-alignment with instruction backtranslation}.
\newblock \emph{CoRR}, abs/2308.06259.

\bibitem[{Li et~al.(2023{\natexlab{b}})Li, Zhang, Dubois, Taori, Gulrajani, Guestrin, Liang, and Hashimoto}]{li2023alpacaeval}
Xuechen Li, Tianyi Zhang, Yann Dubois, Rohan Taori, Ishaan Gulrajani, Carlos Guestrin, Percy Liang, and Tatsunori~B Hashimoto. 2023{\natexlab{b}}.
\newblock Alpacaeval: An automatic evaluator of instruction-following models.
\newblock \emph{GitHub repository}.

\bibitem[{Lin(2004)}]{lin2004rouge}
Chin-Yew Lin. 2004.
\newblock Rouge: A package for automatic evaluation of summaries.
\newblock In \emph{Text summarization branches out}, pages 74--81.

\bibitem[{Longpre et~al.(2023)Longpre, Hou, Vu, Webson, Chung, Tay, Zhou, Le, Zoph, Wei, and Roberts}]{DBLP:conf/icml/LongpreHVWCTZLZ23}
Shayne Longpre, Le~Hou, Tu~Vu, Albert Webson, Hyung~Won Chung, Yi~Tay, Denny Zhou, Quoc~V. Le, Barret Zoph, Jason Wei, and Adam Roberts. 2023.
\newblock \href {https://proceedings.mlr.press/v202/longpre23a.html} {The flan collection: Designing data and methods for effective instruction tuning}.
\newblock In \emph{International Conference on Machine Learning, {ICML} 2023, 23-29 July 2023, Honolulu, Hawaii, {USA}}, volume 202 of \emph{Proceedings of Machine Learning Research}, pages 22631--22648. {PMLR}.

\bibitem[{Mishra et~al.(2022)Mishra, Khashabi, Baral, and Hajishirzi}]{DBLP:conf/acl/MishraKBH22}
Swaroop Mishra, Daniel Khashabi, Chitta Baral, and Hannaneh Hajishirzi. 2022.
\newblock \href {https://doi.org/10.18653/v1/2022.acl-long.244} {Cross-task generalization via natural language crowdsourcing instructions}.
\newblock In \emph{Proceedings of the 60th Annual Meeting of the Association for Computational Linguistics (Volume 1: Long Papers), {ACL} 2022, Dublin, Ireland, May 22-27, 2022}, pages 3470--3487. Association for Computational Linguistics.

\bibitem[{OpenAI(2023)}]{Chatgpt}
OpenAI. 2023.
\newblock \href {https://openai.com/blog/chatgpt/} {Chatgpt}.

\bibitem[{Peng et~al.(2023)Peng, Li, He, Galley, and Gao}]{DBLP:journals/corr/abs-2304-03277}
Baolin Peng, Chunyuan Li, Pengcheng He, Michel Galley, and Jianfeng Gao. 2023.
\newblock \href {https://doi.org/10.48550/arXiv.2304.03277} {Instruction tuning with {GPT-4}}.
\newblock \emph{CoRR}, abs/2304.03277.

\bibitem[{Sanh et~al.(2022)Sanh, Webson, Raffel, Bach, Sutawika, Alyafeai, Chaffin, Stiegler, Raja, Dey, Bari, Xu, Thakker, Sharma, Szczechla, Kim, Chhablani, Nayak, Datta, Chang, Jiang, Wang, Manica, Shen, Yong, Pandey, Bawden, Wang, Neeraj, Rozen, Sharma, Santilli, F{\'{e}}vry, Fries, Teehan, Scao, Biderman, Gao, Wolf, and Rush}]{DBLP:conf/iclr/SanhWRBSACSRDBX22}
Victor Sanh, Albert Webson, Colin Raffel, Stephen~H. Bach, Lintang Sutawika, Zaid Alyafeai, Antoine Chaffin, Arnaud Stiegler, Arun Raja, Manan Dey, M~Saiful Bari, Canwen Xu, Urmish Thakker, Shanya~Sharma Sharma, Eliza Szczechla, Taewoon Kim, Gunjan Chhablani, Nihal~V. Nayak, Debajyoti Datta, Jonathan Chang, Mike~Tian{-}Jian Jiang, Han Wang, Matteo Manica, Sheng Shen, Zheng~Xin Yong, Harshit Pandey, Rachel Bawden, Thomas Wang, Trishala Neeraj, Jos Rozen, Abheesht Sharma, Andrea Santilli, Thibault F{\'{e}}vry, Jason~Alan Fries, Ryan Teehan, Teven~Le Scao, Stella Biderman, Leo Gao, Thomas Wolf, and Alexander~M. Rush. 2022.
\newblock \href {https://openreview.net/forum?id=9Vrb9D0WI4} {Multitask prompted training enables zero-shot task generalization}.
\newblock In \emph{The Tenth International Conference on Learning Representations, {ICLR} 2022, Virtual Event, April 25-29, 2022}. OpenReview.net.

\bibitem[{Taori et~al.(2023)Taori, Gulrajani, Zhang, Dubois, Li, Guestrin, Liang, and Hashimoto}]{alpaca}
Rohan Taori, Ishaan Gulrajani, Tianyi Zhang, Yann Dubois, Xuechen Li, Carlos Guestrin, Percy Liang, and Tatsunori~B. Hashimoto. 2023.
\newblock Stanford alpaca: An instruction-following llama model.
\newblock \url{https://github.com/tatsu-lab/stanford_alpaca}.

\bibitem[{Touvron et~al.(2023{\natexlab{a}})Touvron, Lavril, Izacard, Martinet, Lachaux, Lacroix, Rozi{\`{e}}re, Goyal, Hambro, Azhar, Rodriguez, Joulin, Grave, and Lample}]{DBLP:journals/corr/abs-2302-13971}
Hugo Touvron, Thibaut Lavril, Gautier Izacard, Xavier Martinet, Marie{-}Anne Lachaux, Timoth{\'{e}}e Lacroix, Baptiste Rozi{\`{e}}re, Naman Goyal, Eric Hambro, Faisal Azhar, Aur{\'{e}}lien Rodriguez, Armand Joulin, Edouard Grave, and Guillaume Lample. 2023{\natexlab{a}}.
\newblock \href {https://doi.org/10.48550/arXiv.2302.13971} {Llama: Open and efficient foundation language models}.
\newblock \emph{CoRR}, abs/2302.13971.

\bibitem[{Touvron et~al.(2023{\natexlab{b}})Touvron, Martin, Stone, Albert, Almahairi, Babaei, Bashlykov, Batra, Bhargava, Bhosale, Bikel, Blecher, Canton{-}Ferrer, Chen, Cucurull, Esiobu, Fernandes, Fu, Fu, Fuller, Gao, Goswami, Goyal, Hartshorn, Hosseini, Hou, Inan, Kardas, Kerkez, Khabsa, Kloumann, Korenev, Koura, Lachaux, Lavril, Lee, Liskovich, Lu, Mao, Martinet, Mihaylov, Mishra, Molybog, Nie, Poulton, Reizenstein, Rungta, Saladi, Schelten, Silva, Smith, Subramanian, Tan, Tang, Taylor, Williams, Kuan, Xu, Yan, Zarov, Zhang, Fan, Kambadur, Narang, Rodriguez, Stojnic, Edunov, and Scialom}]{DBLP:journals/corr/abs-2307-09288}
Hugo Touvron, Louis Martin, Kevin Stone, Peter Albert, Amjad Almahairi, Yasmine Babaei, Nikolay Bashlykov, Soumya Batra, Prajjwal Bhargava, Shruti Bhosale, Dan Bikel, Lukas Blecher, Cristian Canton{-}Ferrer, Moya Chen, Guillem Cucurull, David Esiobu, Jude Fernandes, Jeremy Fu, Wenyin Fu, Brian Fuller, Cynthia Gao, Vedanuj Goswami, Naman Goyal, Anthony Hartshorn, Saghar Hosseini, Rui Hou, Hakan Inan, Marcin Kardas, Viktor Kerkez, Madian Khabsa, Isabel Kloumann, Artem Korenev, Punit~Singh Koura, Marie{-}Anne Lachaux, Thibaut Lavril, Jenya Lee, Diana Liskovich, Yinghai Lu, Yuning Mao, Xavier Martinet, Todor Mihaylov, Pushkar Mishra, Igor Molybog, Yixin Nie, Andrew Poulton, Jeremy Reizenstein, Rashi Rungta, Kalyan Saladi, Alan Schelten, Ruan Silva, Eric~Michael Smith, Ranjan Subramanian, Xiaoqing~Ellen Tan, Binh Tang, Ross Taylor, Adina Williams, Jian~Xiang Kuan, Puxin Xu, Zheng Yan, Iliyan Zarov, Yuchen Zhang, Angela Fan, Melanie Kambadur, Sharan Narang, Aur{\'{e}}lien Rodriguez, Robert Stojnic, Sergey Edunov,
  and Thomas Scialom. 2023{\natexlab{b}}.
\newblock \href {https://doi.org/10.48550/arXiv.2307.09288} {Llama 2: Open foundation and fine-tuned chat models}.
\newblock \emph{CoRR}, abs/2307.09288.

\bibitem[{Wang et~al.(2023)Wang, Kordi, Mishra, Liu, Smith, Khashabi, and Hajishirzi}]{DBLP:conf/acl/WangKMLSKH23}
Yizhong Wang, Yeganeh Kordi, Swaroop Mishra, Alisa Liu, Noah~A. Smith, Daniel Khashabi, and Hannaneh Hajishirzi. 2023.
\newblock \href {https://doi.org/10.18653/v1/2023.acl-long.754} {Self-instruct: Aligning language models with self-generated instructions}.
\newblock In \emph{Proceedings of the 61st Annual Meeting of the Association for Computational Linguistics (Volume 1: Long Papers), {ACL} 2023, Toronto, Canada, July 9-14, 2023}, pages 13484--13508. Association for Computational Linguistics.

\bibitem[{Wang et~al.(2022)Wang, Mishra, Alipoormolabashi, Kordi, Mirzaei, Naik, Ashok, Dhanasekaran, Arunkumar, Stap, Pathak, Karamanolakis, Lai, Purohit, Mondal, Anderson, Kuznia, Doshi, Pal, Patel, Moradshahi, Parmar, Purohit, Varshney, Kaza, Verma, Puri, Karia, Doshi, Sampat, Mishra, A, Patro, Dixit, and Shen}]{DBLP:conf/emnlp/WangMAKMNADASPK22}
Yizhong Wang, Swaroop Mishra, Pegah Alipoormolabashi, Yeganeh Kordi, Amirreza Mirzaei, Atharva Naik, Arjun Ashok, Arut~Selvan Dhanasekaran, Anjana Arunkumar, David Stap, Eshaan Pathak, Giannis Karamanolakis, Haizhi~Gary Lai, Ishan Purohit, Ishani Mondal, Jacob Anderson, Kirby Kuznia, Krima Doshi, Kuntal~Kumar Pal, Maitreya Patel, Mehrad Moradshahi, Mihir Parmar, Mirali Purohit, Neeraj Varshney, Phani~Rohitha Kaza, Pulkit Verma, Ravsehaj~Singh Puri, Rushang Karia, Savan Doshi, Shailaja~Keyur Sampat, Siddhartha Mishra, Sujan~Reddy A, Sumanta Patro, Tanay Dixit, and Xudong Shen. 2022.
\newblock \href {https://doi.org/10.18653/v1/2022.emnlp-main.340} {Super-naturalinstructions: Generalization via declarative instructions on 1600+ {NLP} tasks}.
\newblock In \emph{Proceedings of the 2022 Conference on Empirical Methods in Natural Language Processing, {EMNLP} 2022, Abu Dhabi, United Arab Emirates, December 7-11, 2022}, pages 5085--5109. Association for Computational Linguistics.

\bibitem[{Wei et~al.(2022)Wei, Bosma, Zhao, Guu, Yu, Lester, Du, Dai, and Le}]{DBLP:conf/iclr/WeiBZGYLDDL22}
Jason Wei, Maarten Bosma, Vincent~Y. Zhao, Kelvin Guu, Adams~Wei Yu, Brian Lester, Nan Du, Andrew~M. Dai, and Quoc~V. Le. 2022.
\newblock \href {https://openreview.net/forum?id=gEZrGCozdqR} {Finetuned language models are zero-shot learners}.
\newblock In \emph{The Tenth International Conference on Learning Representations, {ICLR} 2022, Virtual Event, April 25-29, 2022}. OpenReview.net.

\bibitem[{Xu et~al.(2023)Xu, Sun, Zheng, Geng, Zhao, Feng, Tao, and Jiang}]{DBLP:journals/corr/abs-2304-12244}
Can Xu, Qingfeng Sun, Kai Zheng, Xiubo Geng, Pu~Zhao, Jiazhan Feng, Chongyang Tao, and Daxin Jiang. 2023.
\newblock \href {https://doi.org/10.48550/arXiv.2304.12244} {Wizardlm: Empowering large language models to follow complex instructions}.
\newblock \emph{CoRR}, abs/2304.12244.

\bibitem[{Yin et~al.(2023)Yin, Liu, Yin, Zhong, Bansal, Han, and Chang}]{DBLP:journals/corr/abs-2305-14327}
Da~Yin, Xiao Liu, Fan Yin, Ming Zhong, Hritik Bansal, Jiawei Han, and Kai{-}Wei Chang. 2023.
\newblock \href {https://doi.org/10.48550/arXiv.2305.14327} {Dynosaur: {A} dynamic growth paradigm for instruction-tuning data curation}.
\newblock \emph{CoRR}, abs/2305.14327.

\bibitem[{Zhang et~al.(2023)Zhang, Press, Merrill, Liu, and Smith}]{DBLP:journals/corr/abs-2305-13534}
Muru Zhang, Ofir Press, William Merrill, Alisa Liu, and Noah~A. Smith. 2023.
\newblock \href {https://doi.org/10.48550/arXiv.2305.13534} {How language model hallucinations can snowball}.
\newblock \emph{CoRR}, abs/2305.13534.

\bibitem[{Zhang et~al.(2019)Zhang, Kishore, Wu, Weinberger, and Artzi}]{zhang2019bertscore}
Tianyi Zhang, Varsha Kishore, Felix Wu, Kilian~Q Weinberger, and Yoav Artzi. 2019.
\newblock Bertscore: Evaluating text generation with bert.
\newblock \emph{arXiv preprint arXiv:1904.09675}.

\bibitem[{Zheng et~al.(2023)Zheng, Huang, and Chang}]{zheng2023does}
Shen Zheng, Jie Huang, and Kevin Chen-Chuan Chang. 2023.
\newblock Why does chatgpt fall short in answering questions faithfully?
\newblock \emph{arXiv preprint arXiv:2304.10513}.

\end{thebibliography}
\bibliographystyle{acl_natbib}

\newpage
\appendix
\section{Full Prompt to Construct the Meta-Training Set}
\subsection{Prompt for Constructing $\Omega_a$}
\label{app:full_prompt_1}
The full prompt for building the meta-training set is as follows.

\texttt{\footnotesize
For the given text, design a task.\\
Each task contains three fields, instruction, input, and output. instruction defines a task in natural language. \\
Instruction is a complete definition of how an input text (e.g., a sentence or a document) is expected to be mapped to an output text.\\
Requiring instruction, input, and output are derived from text wherever possible.\\
Input can be empty to indicate that the task has no input.\\
Instruction must be in imperative sentences formal.\\
Here are demonstrations where your response should be as different from theirs as possible.\\
\{\} \\
\#text\#: "\{\}"}

\subsection{Prompt for Constructing $\Omega_d$}
\label{app:full_prompt_2}
The full prompt for building the meta-training set is as follows.

\texttt{\footnotesize
Combine the following instruction and output into a single coherent text. \\
You can add, delete, or modify some content as appropriate to make the combined text logically sound. \\
\#instruction\#: "\{\}"\\
\#output\#: "\{\}"}

\end{document}